\documentclass[11pt]{article}
\usepackage{emnlp2017}
\usepackage{times}
\usepackage{color}
\usepackage{bm}
\usepackage{latexsym}
\usepackage{amssymb,amsmath}
\usepackage{mathrsfs}
\usepackage{graphicx}
\usepackage{caption}
\usepackage{url}
\usepackage{caption}
\usepackage{tabularx}
\usepackage{booktabs}
\usepackage{mathrsfs}
\usepackage{makecell}
\usepackage{multirow}
\usepackage[linesnumbered,ruled,vlined]{algorithm2e}

\emnlpfinalcopy

\title{Investigating Capsule Networks with Dynamic Routing for \break Text Classification}

\author{Wei Zhao$^{\tt 1,2}$,  Jianbo Ye$^{\tt 3}$,  Min Yang$^{\tt 1\thanks{\quad Corresponding author (min.yang@siat.ac.cn)}}$, Zeyang Lei$^{\tt 4}$,  Soufei Zhang$^{\tt 5}$, Zhou Zhao$^{\tt 6}$\\
{$^{\tt 1}$ Shenzhen Institutes of Advanced Technology, Chinese Academy of Sciences}\\
{$^{\tt 2}$ Tencent} \qquad \qquad {$^{\tt 3}$ Pennsylvania State University}\\
{$^{\tt 4}$ Graduate School at Shenzhen, Tsinghua University }\\
{$^{\tt 5}$ Nanjing University of Posts and Telecommunications} \qquad {$^{\tt 6}$ Zhejiang University} \\
}

\date{}

\begin{document}
\maketitle
\begin{abstract}
In this study, we explore capsule networks with dynamic routing for text classification. We propose three strategies to stabilize the dynamic routing process to alleviate the disturbance of some noise capsules which may contain ``background'' information or have not been successfully trained. A series of experiments are conducted with capsule networks on six text classification benchmarks.  Capsule networks achieve competitive results over the compared baseline methods on 4 out of 6 datasets, which shows the effectiveness of capsule networks for text classification. We additionally show that capsule networks exhibit significant improvement when transfer single-label to multi-label text classification over the competitors. To the best of our knowledge, this is the first work that capsule networks have been empirically investigated for text modeling\footnote{Codes are publicly available at: \url{https://github.com/andyweizhao/capsule_text_classification}.}.
\end{abstract}

\section{Introduction} 
Modeling articles or sentences computationally is a fundamental topic in natural language processing. It could be as simple as 
a keyword/phrase matching problem, but it could also be a nontrivial problem if compositions, hierarchies, 
and structures of texts are considered. For example, a news article which
mentions a single phrase ``US election'' may be categorized into the political news with high probability. 
But it could be very difficult for a computer to predict which presidential candidate is favored by its author,
or whether the author's view in the article is more liberal or more conservative. 

Earlier efforts in modeling texts have achieved
limited success on text categorization using a simple bag-of-words classifier~\cite{joachims1998text,mccallum1998comparison}, 
implying understanding the meaning of the individual word or n-gram is a necessary step towards more sophisticated models. 
It is therefore not a surprise that distributed representations of words, a.k.a. word embeddings, have
received great attention from NLP community addressing the question ``what'' to be modeled at the basic level~\cite{mikolov2013distributed,pennington2014glove}.  
In order to model higher level concepts and facts in texts, an NLP researcher 
has to think cautiously the so-called ``what'' question: \textit{what is actually modeled beyond word meanings}. 
A common approach to the question is to treat the texts as sequences and focus on their spatial patterns, whose representatives include convolutional neural networks (CNNs)~\cite{kim2014convolutional,zhang2015character,conneau2017very} and long short-term memory networks (LSTMs)~\cite{tai2015improved,mousa2017contextual}. 
Another common approach is to completely ignore the order of words but focus on their compositions as a collection,
whose representatives include probabilistic topic modeling~\cite{blei2003latent,mcauliffe2008supervised} 
and Earth Mover's Distance based modeling~\cite{kusner2015word,ye2017determining}. 

Those two approaches, albeit quite different from the computational perspective, actually follow
a common measure to be diagnosed regarding their answers to the ``what'' question. 
In neural network approaches, spatial patterns aggregated at lower levels contribute to representing higher level concepts. 
Here, they form a recursive process to articulate what to be modeled. For example, CNN builds convolutional feature detectors
to extract local patterns from a window of vector sequences and uses max-pooling to select the most prominent ones. 
It then hierarchically builds such pattern extraction pipelines at multiple levels.
Being a spatially sensitive model, CNN pays a price for the inefficiency of replicating feature detectors on a grid. 
As argued in~\cite{sabour2017dynamic}, one has to choose between replicating detectors whose size 
grows exponentially with the number of dimensions, or increasing the volume of the labeled training set in a similar exponential way.  
On the other hand, methods that are spatially insensitive are \textit{perfectly}
efficient at the inference time regardless of any order of words or local patterns. 
However, they are unavoidably more restricted to encode rich structures presented in a sequence.
Improving the efficiency to encode spatial patterns while keeping the flexibility of their representation capability
is thus a central issue. 

A recent method called capsule network introduced by~\newcite{sabour2017dynamic} possesses this attractive potential
to address the aforementioned issue. They introduce an iterative routing process to decide the credit attribution between nodes 
from lower and higher layers. A metaphor (also as an argument) they made is that human visual system intelligently assigns 
parts to wholes at the inference time without hard-coding patterns to be perspective relevant. As an outcome,
their model could encode the intrinsic spatial relationship between a part and a whole constituting viewpoint
invariant knowledge that automatically generalizes to novel viewpoints. 
In our work, we follow a similar spirit
to use this technique in modeling texts. 
Three strategies are proposed to stabilize the dynamic routing process to alleviate the disturbance of some noise capsules which may contain ``background'' information such as stop words and the words that are unrelated to specific categories.
We conduct a series of experiments with capsule networks on  top of the pre-trained word vectors for six text classification benchmarks. More importantly, we show that capsule networks 
achieves significant improvement when transferring single-label to multi-label text classifications over strong baseline methods.

\section{Our Model}
\begin{figure*}
\centering
\includegraphics[ width=6in]{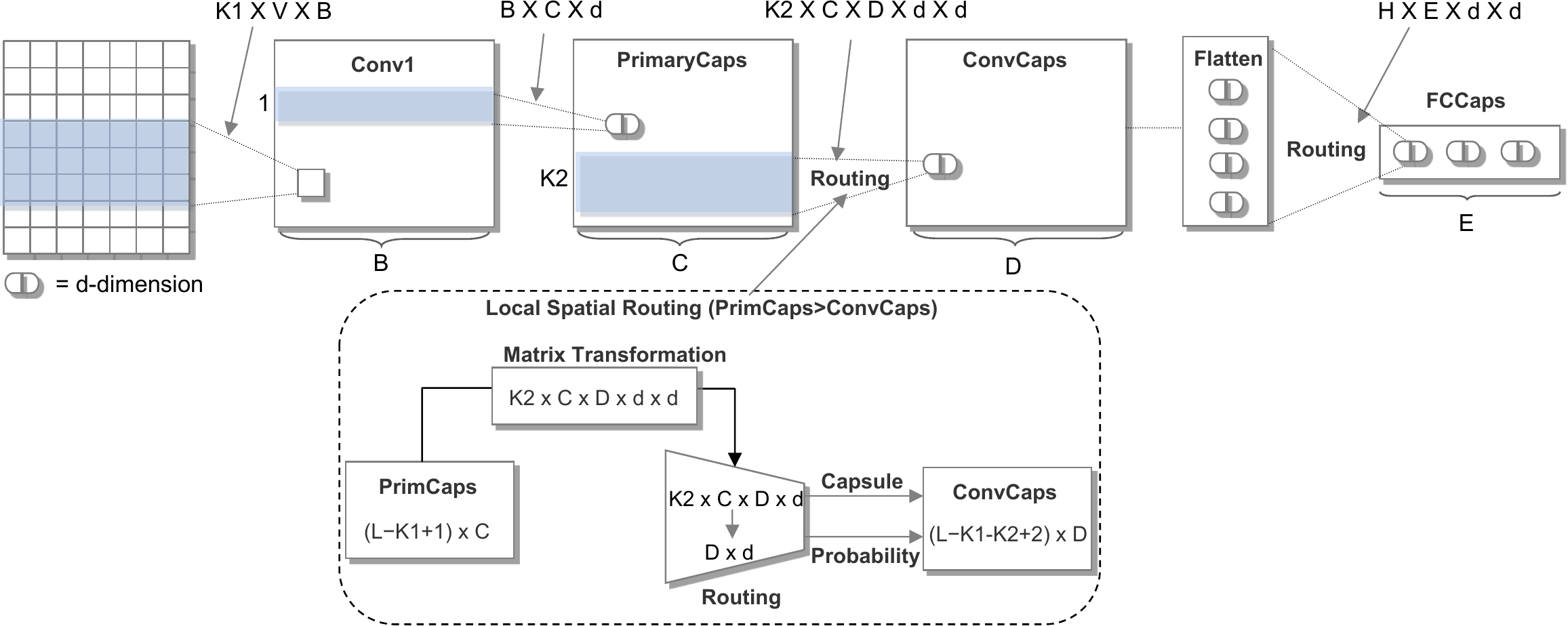}
\caption{The Architecture of Capsule network for text classification. The processes of dynamic routing between consecutive layers are shown in the bottom.}
\label{fig:1}
\vspace{-0.5cm}
\end{figure*}

Our capsule network, depicted in Figure \ref{fig:1}, is a variant of the capsule networks proposed in \newcite{sabour2017dynamic}. It consists of four layers: n-gram convolutional layer, primary capsule layer, convolutional capsule layer, and fully connected capsule layer. In addition, we explore two capsule frameworks to integrate these four components in different ways.  
In the rest of this section, we elaborate the key components in detail.

\subsection{$N$-gram Convolutional Layer}
This layer is a standard convolutional layer which extracts n-gram features at different positions of a sentence through various convolutional filters.

Suppose $\mathbf{x} \in \mathbb{R}^{L\times V}$ denotes the input sentence representation where $L$ is the length of the sentence and $V$ is the embedding size of words. Let $\mathbf{x}_{i}\in \mathbb{R}^V$ be the $V$-dimensional word vector corresponding to the $i$-th word in the sentence. Let $W^a \in \mathbb{R}^{K_{1}\times V}$ be the filter for the convolution operation, where $K_{1}$ is the $N$-gram size while sliding over a sentence for the purpose of detecting features at different positions. 
A filter $W^a$ convolves with the word-window $\mathbf{x}_{i:i+K_{1}-1}$ at each possible position (with stride of 1) to produce a column feature map $\mathbf{m}^{a} \in \mathbb{R}^{L-K_{1}+1}$, each element $m_{i}^a \in \mathbb{R}$ of the feature map is produced by
\begin{equation}
m_{i}^a=f(\mathbf x_{i:i+K_{1}-1}\circ W^a+\mathbf b_{0})
\end{equation}
where $\circ$ is element-wise multiplication, $\mathbf b_{0}$ is a bias term, and $f$ is a nonlinear activate function (i.e., ReLU). We have described the process by which one
feature is extracted from one filter. Hence, for $a=1,\ldots,B$, totally $B$
filters with the same $N$-gram size, one can generate $B$ feature maps which can be rearranged as
\begin{equation}
\mathbf{M}=[\mathbf{m}_{1},\mathbf{m}_{2},...,\mathbf{m}_{\rm{B}}]  \in \mathbb{R}^{(L-K_{1}+1)\times B}
\end{equation}


\subsection{Primary Capsule Layer}
This is the first capsule layer in which the capsules replace the scalar-output feature detectors of CNNs with vector-output capsules to preserve the instantiated parameters such as the local order of words and semantic representations of words.


Suppose $p_{i}\in \mathbb{R}^d$ denotes the instantiated parameters  of a capsule, where $d$ is the dimension of the capsule. Let $W^b \in \mathbb{R}^{B \times d}$ be the filter shared in different sliding windows. For each matrix multiplication, we have a window sliding over each $N$-gram vector denoted as $\mathbf M_{i}\in \mathbb{R}^B$, then the corresponding $N$-gram phrases in the form of capsule are produced with $p_i = (W^b)^T \mathbf M_i$. 

The filter $W^b$ multiplies each $N$-gram vector in $\mathbf {\{M_i\}}_{i=1}^{L-K_1+1}$ with stride of 1 to produce a column-list of capsules $\mathbf{p} \in \mathbb{R}^{(L-K_1+1) \times d }$, each capsule $p_{i} \in \mathbb{R}^d$ in the column-list is computed as 
\begin{equation}
p_{i}= g(W^b \mathbf{M}_i +\mathbf b_{1}) 
\end{equation}
where $g$ is nonlinear squash function through the entire vector, $\mathbf b_{1}$ is the capsule bias term. For all $C$ filters, the generated capsule feature maps can be rearranged as
\begin{equation}
\mathbf{P}=[\mathbf{p}_{1},\mathbf{p}_{2},...,\mathbf{p}_{\rm{C}}] \in \mathbb{R}^{(L-K_1+1)\times C \times d},
\end{equation}
where totally $(L-K_1+1)\times C$ $d$-dimensional vectors are collected as capsules in $\mathbf P$.

\subsubsection{Child-Parent Relationships}
As argued in~\cite{sabour2017dynamic}, capsule network tries to address the representational limitation and exponential inefficiencies of convolutions with transformation matrices. It allows the networks to automatically learn child-parent (or part-whole) relationships. In text classification tasks, different sentences with the same category are supposed to have the similar topic but with different viewpoints.

In this paper, we explore two different types of transformation matrices to generate prediction vector (vote) $\hat{u}_{j|i}\in \mathbb R^d$ from its child capsule $i$ to the parent capsule $j$. The first one shares weights $W^{t_{1}} \in \mathbb{R}^{N\times d\times d}$ across child capsules in the layer below, where $N$ is the number of parent capsules in the layer above. Formally, each corresponding vote can be computed by:
\begin{equation}
\label{eq:1}
\hat{u}_{j|i}=W_j^{t_{1}}u_{i}+\hat{b}_{j|i}\in \mathbb R^d
\end{equation}
where $u_{i}$ is a child-capsule in the layer below and $\hat{b}_{j|i}$ is the capsule bias term. 

In the second design, we replace the shared weight matrix $W_j^{t_{1}}$ with non-shared weight matrix $W_{i,j}^{t_{2}}$,
where the weight matrices $W^{t_2}\in \mathbb{R}^{H \times N\times d\times d}$ and $H$ is the number of child capsules in the layer below.  


\subsection{Dynamic Routing}

The basic idea of dynamic routing is to construct a non-linear map in an iterative manner ensuring that the output of each capsule gets sent to an appropriate parent in the subsequent layer: 
\[\left\{\hat u_{j|i}\in \mathbb R^d\right\}_{i=1,\ldots,H,j=1\ldots,N} \mapsto \left\{v_j\in \mathbb R^d\right\}_{j=1}^N.\]
For each potential parent, the capsule network can increase or decrease the connection strength by dynamic routing, which is more effective than the primitive routing strategies such as max-pooling in CNN that  essentially detects whether a feature is present in any position of the text, but loses spatial information about the feature. 
We explore three strategies to boost the accuracy of routing process by alleviating the disturbance of some noisy capsules:
\paragraph{Orphan Category} Inspired by \newcite{sabour2017dynamic}, an additional ``orphan'' category is added to the network, which can capture the ``background'' information of the text such as stop words and the words that are unrelated to specific categories, helping the capsule network model the child-parent relationship more efficiently. Adding ``orphan'' category in the text is more effective than in image since there is no single consistent ``background'' object in images, while the stop words are consistent in texts such as predicate ``s'', ``am'' and pronouns ``his'', ``she''. 

\paragraph{Leaky-Softmax} We explore Leaky-Softmax \newcite{sabour2017dynamic} in the place of standard softmax while updating connection strength between the children capsules and their parents. 
Despite the orphan category in the last capsule layer, we also need a light-weight method between two consecutive layers to route the noise child capsules to  extra dimension without any additional parameters and computation consuming.

\paragraph{Coefficients Amendment} 
We also attempt to use the probability of existence of child capsules in the layer below to iteratively amend the connection strength as Eq.\ref{eq:2}.

\begin{algorithm}
  \label{alg:1}
  \caption{Dynamic Routing Algorithm}
  \textbf{procedure} ROUTING(\textbf{$\hat{u}_{j|i}$}, $\hat{a}_{j|i}$, $r$, $l$)\\
        Initialize the logits of coupling coefficients $b_{j|i}=0$ \\
  	\For{r \rm{iterations}}
	{
	for all capsule $i$ in layer $l$ and capsule $j$ in layer $l+1$: 
$c_{j|i}=\hat{a}_{j|i} \cdot \text{leaky-softmax}(b_{j|i})$\\
	    for all capsule $j$ in layer $l+1$: $v_{j}=g(\sum_{i}c_{j|i}\hat{u}_{j|i})$, \ \ $a_{j}=|v_{j}|$\\
	    for all capsule $i$ in layer $l$ and capsule $j$ in layer $l+1$: $b_{j|i}=b_{j|i}+\hat{u}_{j|i} \cdot v_{j}$\\
	}
  \textbf{return} $v_{j}$,$a_{j}$\\
\end{algorithm}

Given each prediction vector $\hat{u}_{j|i}$ and its probability of existence $\hat{a}_{j|i}$, where $\hat{a}_{j|i}=\hat{a}_{i}$, each iterative coupling coefficient of connection strength $c_{j|i}$ is updated by
\begin{equation}
\label{eq:2}
c_{j|i}=\hat{a}_{j|i} \cdot \text{leaky-softmax}(b_{j|i})
\end{equation}
where $b_{j|i}$ is the logits of coupling coefficients. Each parent capsule $v_{j}$ in the layer above is a weighted sum over all prediction vectors $\hat{u}_{j|i}$:
\begin{equation}
v_{j}=g(\sum_{i}c_{j|i}\hat{u}_{j|i}),\ \ a_{j}=|v_{j}|
\end{equation}
where $a_{j}$ is the probabilities of parent capsules, $g$ is nonlinear squash function \newcite{sabour2017dynamic} through the entire vector.  Once all of the parent capsules are produced, each coupling coefficient $b_{j|i}$ is updated by:
\begin{equation}
b_{j|i}=b_{j|i}+\hat{u}_{j|i} \cdot v_{j}
\end{equation}
For simplicity of notation, the parent capsules and their probabilities in the layer above are denoted as
\begin{equation}
v, a=\rm{Routing}(\hat{u})
\end{equation}
where $\hat{u}$ denotes all of the child capsules in the layer below, $v$ denotes all of the parent-capsules and their probabilities $a$.

Our dynamic routing algorithm is summarized in Algorithm 1.

\subsection{Convolutional Capsule Layer}
In this layer, each capsule is connected only to a local region $K_2 \times C$ spatially in the layer below. Those capsules in the region multiply transformation matrices to learn child-parent relationships followed by routing by agreement to produce parent capsules in the layer above. 


Suppose $W^{c_{1}}\in\mathbb{R}^{D\times d\times d}$ and $W^{c_{2}} \in \mathbb{R}^{K_2 \times C\times D\times d\times d}$ denote shared and non-shared weights, respectively, where $K_2\cdot C$ is the number of child capsules in a local region in the layer below, $D$ is the number of parent capsules which the child capsules are sent to. When the transformation matrices are shared across the child capsules, each potential parent-capsule $\hat{u}_{j|i}$ is produced by
\begin{equation}
\label{eq:3}
\hat{u}_{j|i}=W_{j}^{c_{1}}u_{i}+\hat{b}_{j|i}
\end{equation}
where $\hat{b}_{j|i}$ is the capsule bias term, $u_{i}$ is a child capsule in a local region $K_2 \times C$ and $W_j^{c_1} $ is the $j_{\rm{th}}$ matrix in tensor $W^{c_{1}}$. Then, we use routing-by-agreement to produce parent capsules feature maps totally $(L-K_1-K_2+2)\times D $ $d$-dimensional capsules in this layer.
When using the non-shared weights across the child capsules, we replace the transformation matrix $W_j^{c_{1}}$ in Eq. (\ref{eq:3}) with $W_j^{c_{2}}$.

\subsection{Fully Connected Capsule Layer}
The capsules in the layer below are flattened into a list of capsules and fed into fully connected capsule layer in which capsules are multiplied by transformation matrix $W^{d_{1}} \in \mathbb{R}^{E \times d \times d}$ or $W^{d_{2}} \in \mathbb{R}^{H\times E \times d \times d}$ followed by routing-by-agreement to produce final capsule $v_{j}\in \mathbb{R}^{d}$ and its probability $a_{j} \in \mathbb{R}$ for each category. Here, $H$ is the number of child capsules in the layer below, $E$ is the number of categories plus an extra orphan category.


\begin{figure}
\begin{minipage}{0.3\linewidth}  
	\centerline{\includegraphics[width=2.05cm]{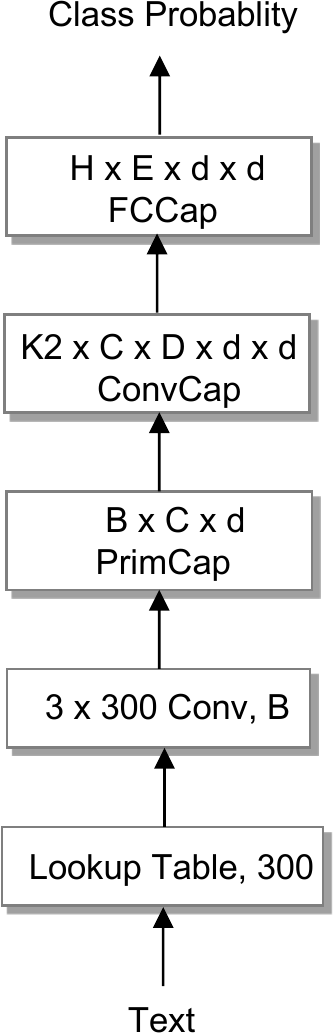}}  
	\centerline{Capsule-A}  
\end{minipage} 
\begin{minipage}{0.68\linewidth}  
	\centerline{\includegraphics[width=5.3cm]{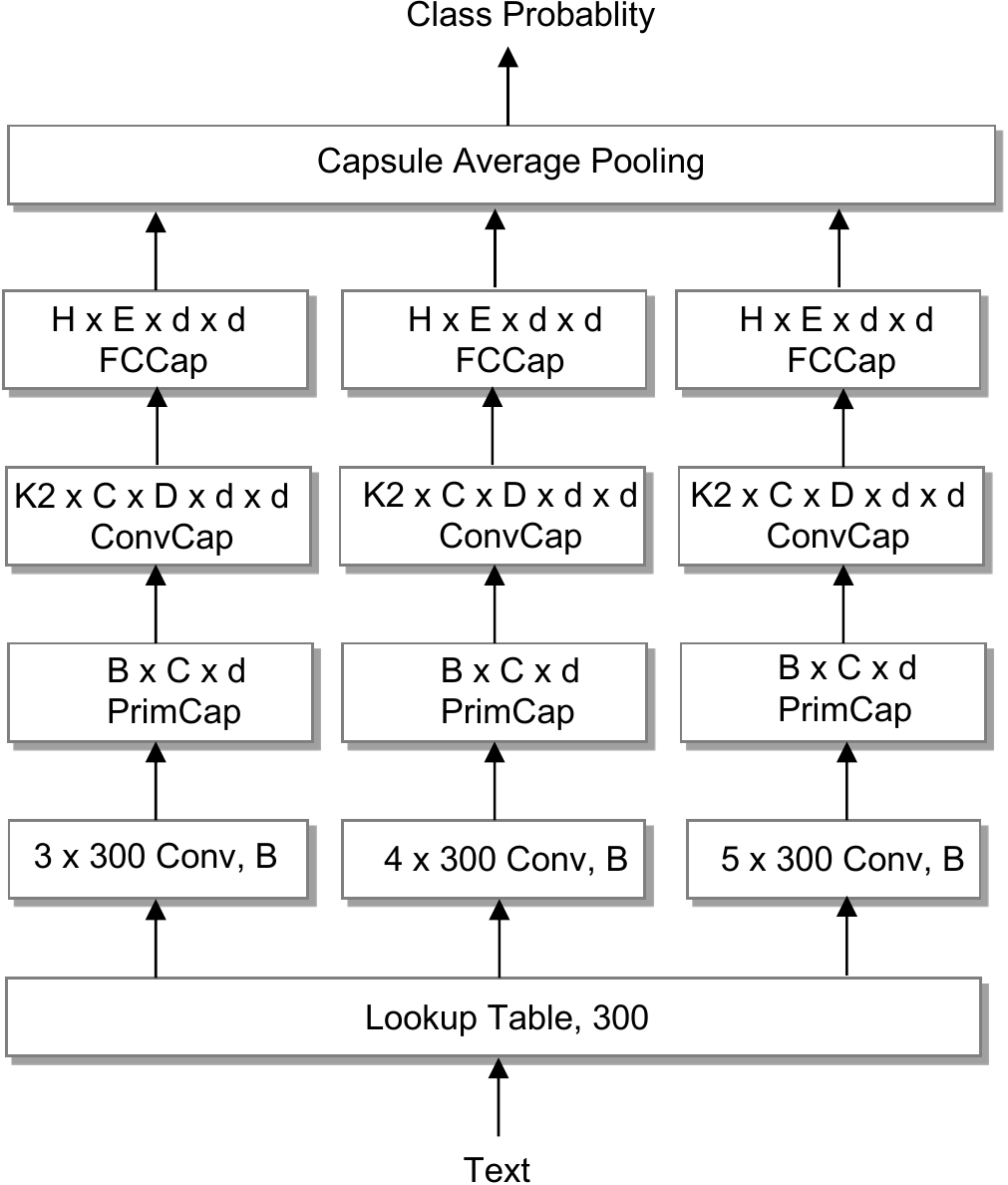}}  
	\centerline{Capsule-B}  
\end{minipage}  
\caption{Two architectures of capsule networks.}  
\label{fig:schema}
\vspace{-0.5cm}
\end{figure} 


\subsection{The Architectures of Capsule Network}
We explore two capsule architectures (denoted as Capsule-A and Capsule-B) to integrate these four components in different ways, as depicted in Figure \ref{fig:schema}. 

Capsule-A starts with an embedding layer which transforms each word in the corpus to a 300-dimensional ($V=300$) word vector, followed by a 3-gram ($K_{1}=3$) convolutional layer with 32 filters ($B=32$) and  a stride of 1 with ReLU non-linearity.  All the other layers are capsule layers starting with a $B\times d$ primary capsule layer with 32 filters ($C=32$), followed by a $3\times C\times d \times d$ ($K_{2}=3$) convolutional capsule layer with 16 filters ($D=16$) and a fully connected capsule layer in sequence. 

Each capsule has 16-dimensional ($d=16$) instantiated parameters and their length (norm) can describe the probability of the existence of capsules. 
The capsule layers are connected by the transformation matrices, and each connection is also multiplied by a routing coefficient that is dynamically computed by routing by agreement mechanism. 


The basic structure of Capsule-B is similar to Capsule-A except that we adopt three parallel networks with filter
windows ($N$) of 3, 4, 5 in the $N$-gram convolutional layer (see Figure \ref{fig:schema}). The final output of the fully connected capsule layer is fed into the average pooling to produce the final results. In this way, Capsule-B can learn more meaningful and comprehensive text representation. 


\section{Experimental Setup}
\subsection{Experimental Datasets}

In order to evaluate the effectiveness of our model, 
we conduct a series of experiments on six benchmarks including: movie reviews (MR) \cite{pang2005seeing}, Stanford Sentiment Treebankan extension of MR (SST-2) \cite{socher2013recursive}, Subjectivity dataset (Subj) \cite{pang2004sentimental},  TREC question dataset (TREC) \cite{li2002learning}, customer review (CR) \cite{hu2004mining}, and AG's news corpus \cite{conneau2017very}. 
These benchmarks cover several text classification tasks such as sentiment classification, question categorization, news categorization. The detailed statistics are presented in Table \ref{tab:0}. 
\begin{table}[h]
\centering
	\resizebox{1\columnwidth}{!}
	{\footnotesize
	\begin{tabular}{|l|lllc|l|}
		\toprule
		Dataset & Train & Dev &Test & Classes & Classification Task \\
		\midrule
		MR &  8.6k & 0.9k & 1.1k & 2 & review classification\\
		SST-2 & 8.6k  & 0.9k & 1.8k & 2 & sentiment analysis\\
		Subj & 8.1k  & 0.9k & 1.0k & 2 & opinion classification\\
		TREC & 5.4k & 0.5k & 0.5k & 6 & question categorization\\
		CR & 3.1k & 0.3k & 0.4k & 2 & review classification\\
		AG's news & 108k & 12.0k & 7.6k & 4 & news categorization\\
		\bottomrule
	\end{tabular}
	}
\caption{Characteristics of the datasets.}
\label{tab:0}
\vspace{-0.3cm}
\end{table}







\subsection{Implementation Details}
In the experiments, we use 300-dimensional word2vec \cite{mikolov2013distributed} vectors to initialize embedding vectors.
We conduct mini-batch with size 50 for AG's news and size 25 for other datasets. We use Adam optimization algorithm with 1e-3 learning rate to train the model. We use 3 iteration of routing for all datasets since it optimizes the loss faster and converges to a lower loss at the end.

\subsection{Baseline methods}
In the experiments, we evaluate and compare our model with several strong baseline methods including: LSTM/Bi-LSTM  \cite{cho2014learning}, tree-structured LSTM (Tree-LSTM) \cite{tai2015improved}, LSTM regularized by linguistic knowledge (LR-LSTM) \cite{qian2016linguistically}, CNN-rand/CNN-static/CNN-non-static \cite{kim2014convolutional}, very deep convolutional network (VD-CNN) \cite{conneau2017very}, and character-level convolutional network (CL-CNN) \cite{zhang2015character}.

\section{Experimental Results}
\begin{table}[h!]
	\centering
	\resizebox{1\columnwidth}{!}{
	
	\begin{tabular}{|l|c|c|c|c|c|c|c}
		\toprule
		 & \textbf{MR} &  \textbf{SST2}& \textbf{Subj} & \textbf{TREC}  & \textbf{CR} & \textbf{AG's} \\
		\midrule
		LSTM  & 75.9 & 80.6 &  89.3 & 86.8  &  78.4 & 86.1  \\
		BiLSTM  & 79.3 & 83.2 &  90.5 & 89.6  &  82.1 & 88.2  \\
		Tree-LSTM  & 80.7 & 85.7 &  91.3 & 91.8   &  83.2 & 90.1 \\
		LR-LSTM  & 81.5 & \textbf{87.5} &  89.9 &  -  &  82.5 & -  \\
		\midrule
		CNN-rand & 76.1 & 82.7 & 89.6 &  91.2  &  79.8 & 92.2 \\
		CNN-static & 81.0 & 86.8 & 93.0 &  92.8  &  84.7 & 91.4 \\
		CNN-non-static  & 81.5 & 87.2 &  93.4 &  \textbf{93.6}  &  84.3 & 92.3\\
		CL-CNN  & - & - &  88.4 &  85.7  &  - & 92.3 \\
		VD-CNN   & - & - &  88.2 &  85.4  &  - & 91.3  \\
		\midrule
		Capsule-A & 81.3 & 86.4 & 93.3  & 91.8 &  83.8 & 92.1\\
		Capsule-B & \textbf{82.3} &86.8 &  \textbf{93.8} & 92.8 &  \textbf{85.1} & \textbf{92.6} \\
                \midrule		
		\bottomrule
	\end{tabular}
	}
    \caption{Comparisons of our capsule networks and baselines on six text classification benchmarks. }
    \label{tab:1}
    \vspace{-0.5cm}
\end{table}

\begin{table}[h!]
\scriptsize
\centering
	\resizebox{1\columnwidth}{!}
	{\footnotesize
	\begin{tabular}{|l|llc|l|}
		\toprule
		Dataset & Train & Dev & Test & Description\\
		\midrule
		Reuters-Multi-label & 5.8k & 0.6k & 0.3k & only multi-label data in test\\
		Reuters-Full & 5.8k  & 0.6k & 3.4k & full data in test\\
		\bottomrule
	\end{tabular}
	}
\caption{Characteristics of Reuters-21578 corpus.}
\label{tab:4}
\vspace{-0.5cm}
\end{table}

\begin{table*}[h!]
\scriptsize
	\centering
    	\resizebox{1.8\columnwidth}{!}{
	
	\begin{tabular}{l |c c c c | c c c c c c c c|}
		\toprule
         &\multicolumn{4}{c|}{Reuters-Multi-label} 
         &\multicolumn{4}{c}{Reuters-Full}
         \\
\cmidrule{2-5}\cmidrule{6-9}
&ER&Precision&Recall &F1&ER&Precision&Recall &F1\\
\midrule
LSTM
&  23.3 &  86.7  &  54.7 &  63.5
&  62.5 &  78.6  &  72.6 &  74.0\\
BiLSTM 
&  26.4 &  82.3  &  55.9 &  64.6
&  65.8 &  83.7  &  75.4 &  77.8\\
\midrule
CNN-rand 
&  22.5 &  88.6  &  56.4 &  67.1
&  63.4 &  78.7  &  71.5 &  73.6\\
CNN-static 
& 27.1  &  91.1  &  59.1 &  69.7 
&  63.3 &  78.5  &  71.2 &  73.3\\
CNN-non-static 
& 27.4 &  92.0  &  59.7 &  70.4 
& 64.1 &  80.6  & 72.7 &  75.0\\

\midrule
Capsule-A 
& 57.2 &  88.2  & 80.1  &  82.0  
& 66.0 &  83.9  &  \textbf{80.5} &  80.2\\
Capsule-B  
&  \textbf{60.3} &  \textbf{95.4}  &  \textbf{82.0} &  \textbf{85.8}  
&  \textbf{67.7} &  \textbf{86.4}  &  80.1 &  \textbf{81.4}\\

	\bottomrule
	\end{tabular}
	}
    \caption{Comparisons of the capability for transferring from single-label to multi-label text classification on Reuters-Multi-label and Reuters-Full datasets.  For fair comparison, we use margin-loss for our model and other baselines.
    }
    \label{tab:5}
    \vspace{-0.5cm}
\end{table*}

\subsection{Quantitative Evaluation}
In our experiments, the evaluation metric is classification accuracy. We summarize the experimental results in Table \ref{tab:1}.
From the results, we observe that the capsule networks achieve best results on 4 out of 6 benchmarks, which verifies the effectiveness of the capsule networks.  
In particular, our model substantially and consistently outperforms the simple deep neural networks such as LSTM, Bi-LSTM and CNN-rand by a noticeable margin on all the experimental datasets. 
Capsule network also achieves competitive results against the more sophisticated deep learning models such as LR-LSTM, Tree-LSTM, VC-CNN and CL-CNN.  
Note that Capsule-B  consistently performs better than Capsule-A since Capsule-B allows to learn more meaningful and comprehensive text representation. For example, a combination of N-gram convolutional layer with filter windows of \{3,4,5\} can capture the 3/4/5-gram features of the text which play a crucial role in text modeling.

\subsection{Ablation Study}
To analyze the effect of varying different components of our capsule architecture for text classification, we also report the ablation test of the capsule-B model in terms of using different setups of the capsule network. 
The experimental results are summarized in Table \ref{tab:ablation}. 
Generally, all three proposed dynamic routing strategies contribute to the effectiveness of Capsule-B by alleviating the disturbance of some noise capsules which may contain ``background'' information such as stop words and the words that are unrelated to specific categories. 
More comprehensive comparison results are demonstrated in Table \ref{tab:a-4} in Supplementary Material.

\begin{table}[!ht]
	\centering
	\resizebox{1\columnwidth}{!}{
	\begin{tabular}{|l |c c |}
		\toprule
         & \textbf{Iteration} & \textbf{Accuracy}\\ 
         \midrule        
Capsule-B + Sabour's routing& 3 & 81.4 \\
 \midrule     
Capsule-B + our routing& 1 & 81.4\\
Capsule-B + our routing& 3 &82.3 \\ 
Capsule-B + our routing& 5 & 81.6\\ 
w/o Leaky-softmax & 3 &81.7 \\ 
w/o Orphan Category& 3 & 81.9 \\ 
w/o Amendent Coeffient & 3 & 82.1 \\
		\bottomrule
	\end{tabular}
	}
    \caption{Ablation study of Capsule-B on MR dataset. The standard routing is routing-by-agreement algorithm without leaky-softmax and orphan category in the last capsule layer. More ablations are discussed in Appendix.}\label{tab:ablation}
    \vspace{-0.3cm}
\end{table}

\begin{table*}[ht!]
	\centering
	\resizebox{2.1\columnwidth}{!}{
	\begin{tabular}{m{12cm}|c|c}
		\toprule
        \hline
		 \textbf{U.K. MONEY RATES FIRM ON LAWSON STERLING TARGETS} & \textbf{Interest Rates}&\textbf{Money/Foreign Exchange}\\
         \hline \textcolor{red}{Interest rates} on the London money market
  were slightly firmer on news U.K. Chancellor of the Exchequer Nigel Lawson had stated \textcolor{red}{target rates} for sterling against the dollar and mark, dealers said.
      They said this had come as a surprise and expected the targets, 2.90 marks and 1.60 dlrs, to be promptly tested in the \textcolor{red}{foreign exchange markets}. Sterling opened 0.3 points lower in trade weighted terms at 71.3.
      Dealers noted the chancellor said he would achieve his goals on sterling by a combination of intervention in \textcolor{red}{currency markets and interest rates}.
      Operators feel the \textcolor{red}{foreign exchanges} are likely to test sterling on the downside and that this seems to make a fall in U.K. Base lending rates even less likely in the near term, dealers said.
      The feeling remains in the market, however, that fundamental factors have not really changed and that a rise in U.K. \textcolor{red}{Interest rates} is not very likely. The market is expected to continue at around these levels, reflecting the current 10 pct base rate level, for some time.
      The key three months interbank rate was 1/16 point firmer at 10 9-7/8 pct.
	&\makecell{\\ \includegraphics[width=4.5in]{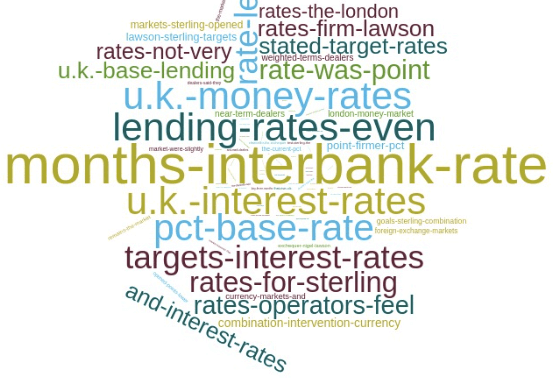} }
	&\makecell{\\ \includegraphics[width=3.3in]{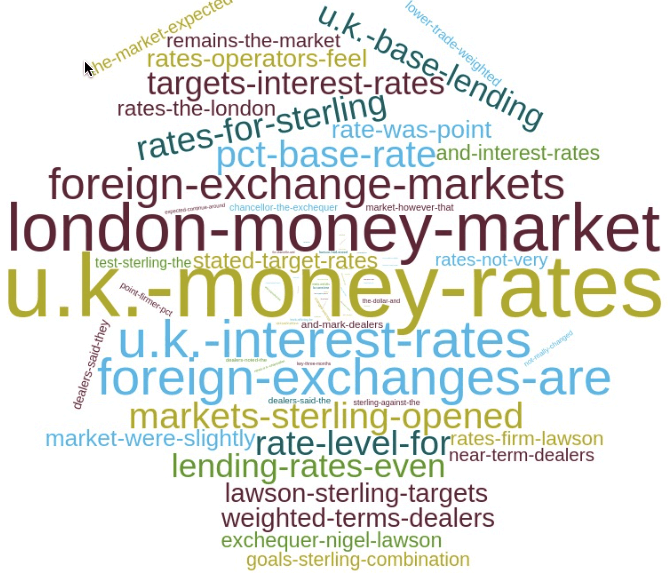} }
    \\
         \hline  
         \multicolumn{3}{l}\makecell{\includegraphics[width=13in]{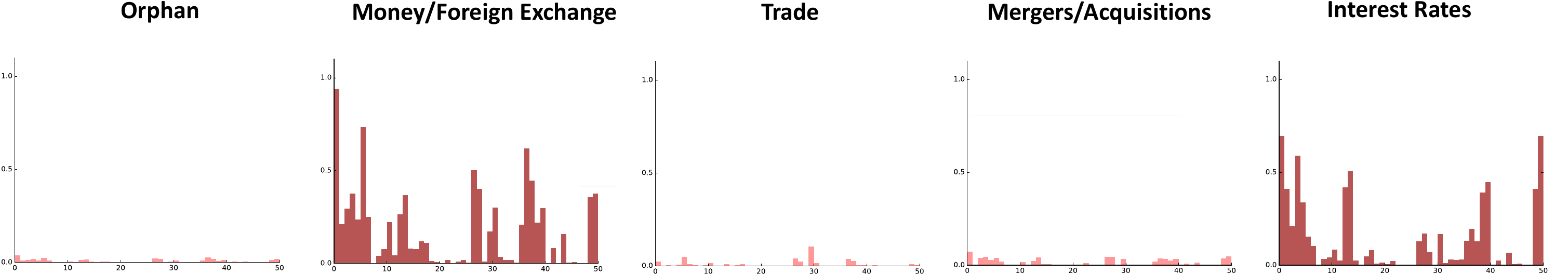} }\\
		\bottomrule   
         \hline
		\bottomrule
	\end{tabular}
	}  
    \caption{Visualization of connection strength between primary capsules and the FC capsules by 3-gram phrases cloud and histogram of the their intensities. x axis denotes primary capsules (3-gram phrases) selected for demonstration, y axis denotes intensity of connection strength. The results are retrieved from Capsule-B trained with 3 routing iterations. The category-specific key-phrases in red color in raw text (first column) are annotated manually for reference.}
     \label{tab:6}
     \vspace{-0.5cm}
\end{table*}



\section{Single-Label to Multi-Label Text Classification}
Capsule network demonstrates promising performance in single-label text classification which assigns a label from a predefined set to a text  (see Table  \ref{tab:1}).  Multi-label text classification is, however, a more challenging practical problem.
From single-label to multi-label (with $n$ category labels) text classification, the label space is expanded from $n$ to $2^n$, thus more training is required to cover the whole label space. For single-label texts, it is practically easy to collect and annotate the samples. However, the burden of collection and annotation for a large scale multi-label text dataset is generally extremely high. 
How deep neural networks (e.g., CNN and LSTM) best cope with multi-label text classification still remains a problem since obtaining large scale of multi-label dataset is a time-consuming and expensive process. 
In this section, we investigate the capability of capsule network on multi-label text classification by using only the single-label samples as training data.
With feature property as part of the information extracted by capsules, we may generalize the model better to multi-label text classification without an over extensive amount of labeled data.  

The evaluation is carried on the Reuters-21578 dataset \cite{LEWIS92b}. This dataset consists of 10,788 documents from the Reuters financial newswire service, where each document contains either multiple labels or a single label. We re-process the corpus to evaluate the capability of capsule networks of transferring from single-label to multi-label text classification. 
For dev and training, we only use the single-label documents in the Reuters dev and training sets. For testing, Reuters-Multi-label only uses the multi-label documents in testing dataset, while Reuters-Full includes all documents in test set.
The characteristics of these two datasets are described in Table \ref{tab:4}.

Following \cite{sorower2010literature}, we adopt Micro Averaged Precision (Precision), Micro Averaged Recall (Recall) and Micro Averaged F1 scores (F1) as the evaluation metrics for multi-label text classification. Any of these scores are firstly computed on individual class labels 
and then averaged over all classes, called label-based measures. 
In addition, we also measure the Exact Match Ratio (ER) which considers partially correct prediction as incorrect and only counts fully correct samples.

The experimental results are summarized in Table \ref{tab:5}. 
From the results, we can observe that the capsule networks 
have substantial and significant improvement in terms of all four evaluation metrics  over the strong baseline methods on the test sets in both Reuters-Multi-label and Reuters-Full datasets. In particular, larger improvement is achieved on Reuters-Multi-label dataset which only contains the multi-label documents in the test set. 
This is within our expectation since the capsule network is capable of preserving the instantiated parameters of the categories trained by single-label documents. The capsule network has much stronger transferring capability than the conventional deep neural networks. 
In addition, the good results on Reuters-Full also indicate that the capsule network has robust superiority over competitors on single-label documents.

\subsection{Connection Strength Visualization}
To visualize the connection strength between capsule layers clearly, we remove the convolutional capsule layer and make the primary capsule layer followed by the fully connected capsule layer directly, where the primary capsules denote N-gram phrases in the form of capsules. The connection strength shows the importance of each primary capsule for text  categories, acting like a parallel attention mechanism. This should allow the capsule networks to recognize multiple categories in the text even though the model is trained on single-label documents. 

Due to space reasons, we choose a multi-label document from Reuters-Multi-label test set whose category labels (i.e., Interest Rates and Money/Foreign Exchange) are correctly predicted (fully correct) by our model with high confidence ($p>0.8$) to report in Table \ref{tab:6}. 
The category-specific phrases such as ``interest rates'' and ``foreign exchange'' are highlighted with red color. We use the tag cloud to visualize the 3-gram phrases for 
Interest Rates and Money/Foreign Exchange categories. The stronger the connection strength, the bigger the font size.  From the results, we observe that capsule networks can correctly recognize and cluster the important phrases with respect to the text categories.
The histograms are used to show the intensity of connection strengths between primary capsules and the fully connected capsules, as shown in Table \ref{tab:6} (bottom line). Due to space reasons, five histograms are demonstrated.
The routing procedure correctly routes the votes into the Interest Rates and Money/Foreign Exchange categories. More examples can be found in Table \ref{tab:a-1}-\ref{tab:a-2} in Supplementary Material.

\begin{figure}[h!]
\centering
\includegraphics[width=2.8in]{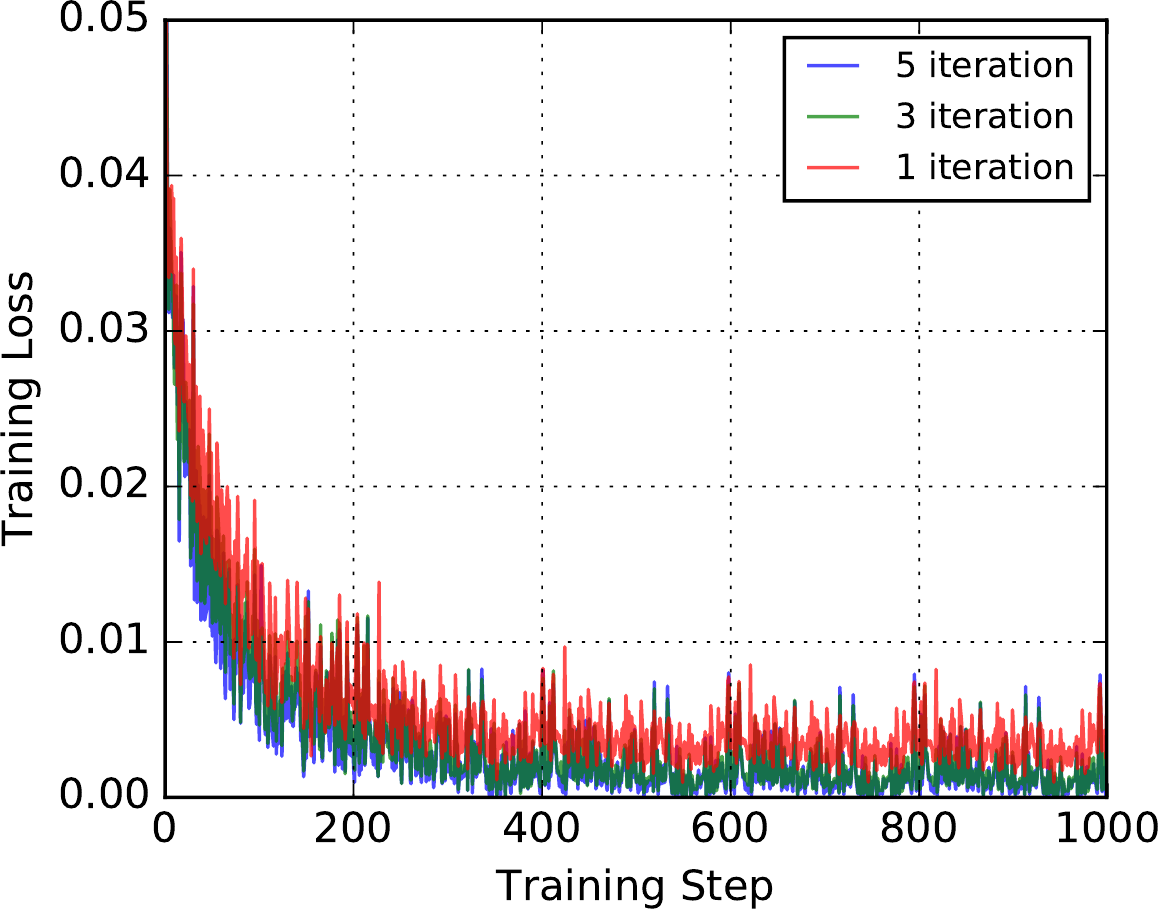}
\caption{Training loss of Capsule-B on Reuters-Multi-label dataset.}
\label{fig:3}
\vspace{-0.5cm}
\end{figure}

To experimentally verify the convergence of the routing algorithm, we also plot learning curve to show the training loss over time with different iterations of routing. From Figure \ref{fig:3}, we observe that the Capsule-B with 3 or 5 iterations of routing optimizes the loss faster and converges to a lower loss at the end than  the capsule network with 1 iteration. 

\section{Related Work} 
Early methods for text classification adopted the typical features such as bag-of-words, n-grams, and their TF-IDF features \cite{zhang2008tfidf} as input of  machine learning algorithms such as support vector machine (SVM) \cite{joachims1998text}, logistic regression \cite{genkin2007large}, naive Bayes (NB) \cite{mccallum1998comparison} for classification. However, these models usually heavily relied on laborious feature engineering or massive extra linguistic resources.

Recent advances in deep neural networks and representation learning have substantially improved the performance of text classification tasks. 
The dominant approaches are recurrent neural networks, in particular LSTMs and CNNs. 
\cite{kim2014convolutional} reported on a series of experiments with CNNs trained on top of pre-trained word vectors for sentence-level classification tasks. The CNN models improved upon the state of the art on 4 out of 7 tasks. 
\cite{zhang2015character} offered an empirical exploration on the use of character-level convolutional networks (Convnets) for text classification and  the experiments showed that Convnets outperformed the traditional models. 
\cite{joulin2016bag} proposed a simple and efficient text classification method fastText, which could be trained on a billion words within ten minutes. 
\cite{conneau2017very} proposed a very deep convolutional networks (with 29 convolutional layers) for text classification. 
\cite{tai2015improved} generalized the LSTM to the  tree-structured network topologies (Tree-LSTM) that achieved best results on two text classification tasks. 

Recently, a novel type of neural network is proposed using the concept of capsules to improve the representational limitations of CNN and RNN. 
\newcite{hinton2011transforming}  firstly introduced the concept of ``capsules'' to address the representational limitations of CNNs and RNNs. Capsules with transformation matrices allowed networks to automatically learn part-whole relationships. 
Consequently, \newcite{sabour2017dynamic} proposed  capsule networks that replaced the scalar-output feature detectors of CNNs with vector-output capsules and max-pooling with routing-by-agreement. The capsule network has shown its potential by achieving a state-of-the-art result on MNIST data. 
Unlike max-pooling in CNN, however, Capsule network do not throw away information about the precise position of the entity within the region. For low－level capsules, location information is “place-coded” by which capsule is active.
\cite{xi2017capsule} further tested out the application of capsule networks on CIFAR data with higher dimensionality.
\cite{hinton2018capsule} proposed a new iterative routing procedure between capsule layers based on the EM algorithm, which achieves significantly better accuracy on the smallNORB data set. 
\cite{zhang2018fast} generalized existing routing methods within the framework of weighted kernel density estimation.
To date, no work investigates the performance of capsule networks in NLP tasks. This study herein takes the lead in this topic.

\section{Conclusion}
In this paper, we investigated capsule networks with dynamic routing for text classification. Three strategies were proposed to boost the performance of the dynamic routing process to alleviate the disturbance of noisy capsules. Extensive experiments on six text classification benchmarks show the effectiveness of capsule networks in text classification. More importantly, capsule networks also show significant improvement when transferring single-label to multi-label text classifications over strong baseline methods. 

\bibliographystyle{emnlp_natbib}
\bibliography{capsule}

\clearpage

\onecolumn
\section*{Supplementary Material}
\setcounter{table}{0}
\renewcommand\thetable{A.\arabic{table}}

To better demonstrate the orphan and other categories with top unigrams, we remove the convolutional capsule layer and make the primary capsule layer followed by the fully connected capsule layer directly, similar to the settings in section 5.1. 
Here, the primary capsules denote uni-grams in the form of capsules. We picked top-20 uni-gram (words) from four categories (i.e., Orphan category, Trade category, Money Exchange category and Interest Rates category) sorted by their connection strengths.

\begin{table}[h!]
	\centering
	\resizebox{0.9\columnwidth}{!}{
	\begin{tabular}{c| l | l |l |l}
		\toprule
		 \textbf{Index}& \textbf{Orphan } &\textbf{Trade } &\textbf{Money Exchange}& \textbf{Interest Rates}\\
		\midrule
	1	& the & trade &money &  Fed (Federal Reserve) \\
	2	& and & Fed (Federal Reserve) &market &  rate \\
	3	& said & market & bank &  pct (Percent of Total) \\
	4	& for & rate & currency &  bank \\
	5	& its & deficit & STG (Sterling) &  market \\
    6	& U.S. & surplus & rate &  repurchase \\
    7	& that & pct (Percent of Total) & repurchase &customer \\
    8	& from & minister & reserves &  federal \\
    9	& mln(Millon) & customer & dollar &  dealers \\
	10	& was & export &customer &  reserve \\
	11	& with & mln(Millon)  & bills &  economists \\
	12	& billion & bank & funds &  Bundesbank \\
	13	& gulf & imports & exchange &  interest \\
	14	& not & money & liquidity &  discount \\
	15	& today & oil & dealers &  trading \\
	
	16	& will & agreements & monetary &  money \\
	17	& they & repurchase & treasury &  lending \\
	18	& had & goods & sterling &  treasury \\
	19	& were & bills & Bundesbank &  bankers \\
	20	& would & shipping & deposits &  agreements \\	
                \midrule		
		\bottomrule
	\end{tabular}
	}
    \caption{Top-20 words picked from four categories (i.e., Orphan category, Trade category, Money Exchange category and Interest Rates category) sorted by their connection strengths.}
    \label{tab:a-3}
\end{table} 

\begin{table}[h]
	\centering
	\resizebox{1\columnwidth}{!}{
	\begin{tabular}{m{15cm}|c|c}
		\toprule
        \hline
		 \textbf{G-7 ISSUES STATEMENT AFTER MEETING} & \textbf{Trade} &\textbf{Money/Foreign Exchange}\\
         \hline Following is the text of a statement
  by the Group of Seven -- the U.S., Japan, West Germany, France,
  Britain, Italy and Canada -- issued after a Washington meeting
  yesterday.
      1. The finance ministers and central bank governors of
  seven major industrial countries met today.
      They continued the process of multilateral surveillance of
  their economies pursuant to the arrangements for strengthened
  economic policy coordination agreed at the 1986 Tokyo summit of
  their heads of state or government.
      2. The ministers and governors reaffirmed the commitment to
  the cooperative approach agreed at the recent Paris meeting,
  and noted the progress achieved in implementing the
  undertakings embodied in the Louvre Agreement.
      In this connection they welcomed the proposals just
  announced by the governing Liberal Democratic Party in Japan
  for extraordinary and urgent measures to stimulate Japan's
  economy through early implementation of a large supplementary
  budget exceeding those of previous years, as well as
  unprecedented front-end loading of public works expenditures.
      They concluded that present and prospective progress in
  implementing the policy undertakings at the Louvre and in this
  statement provided a basis for continuing close cooperation to
  foster the stability of \textcolor{red}{exchange rates}.
	&\makecell{\\ \includegraphics[width=3.5in]{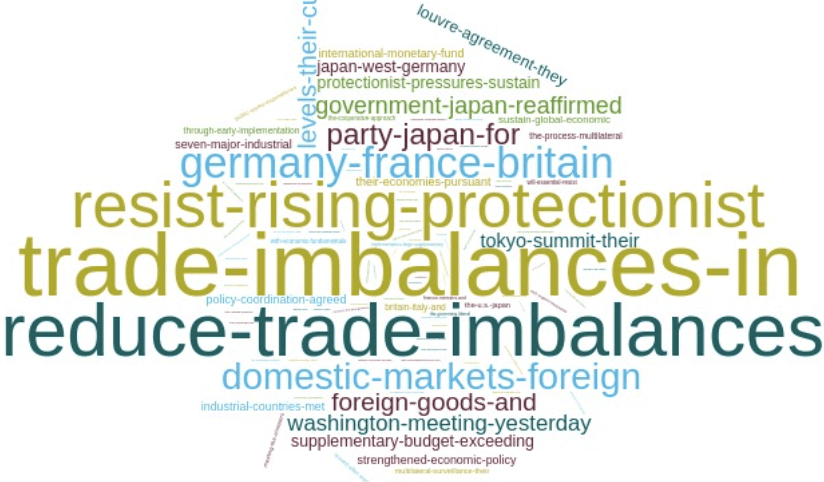} }
	&\makecell{\\ \includegraphics[width=3.5in]{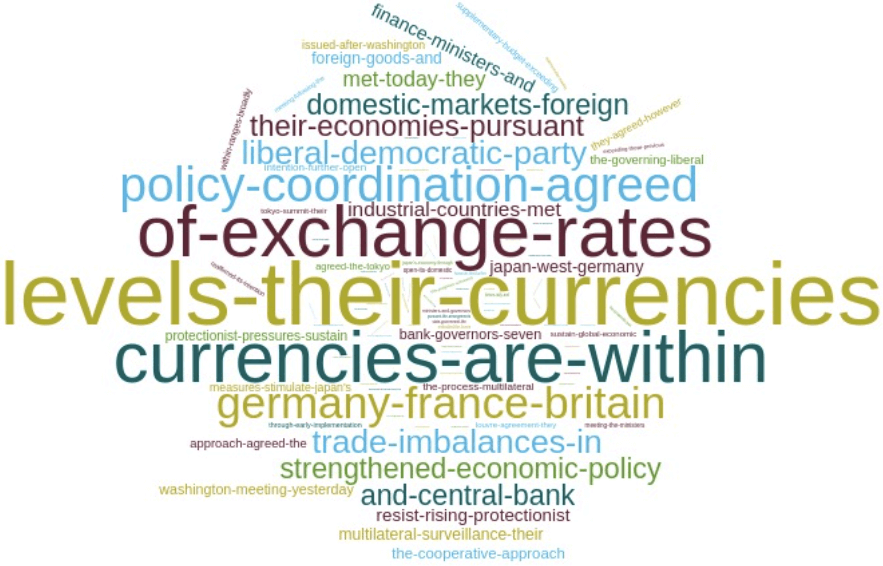} }
    \\
         \bottomrule        
      	 \multicolumn{3}{l}\makecell{\includegraphics[width=13in]{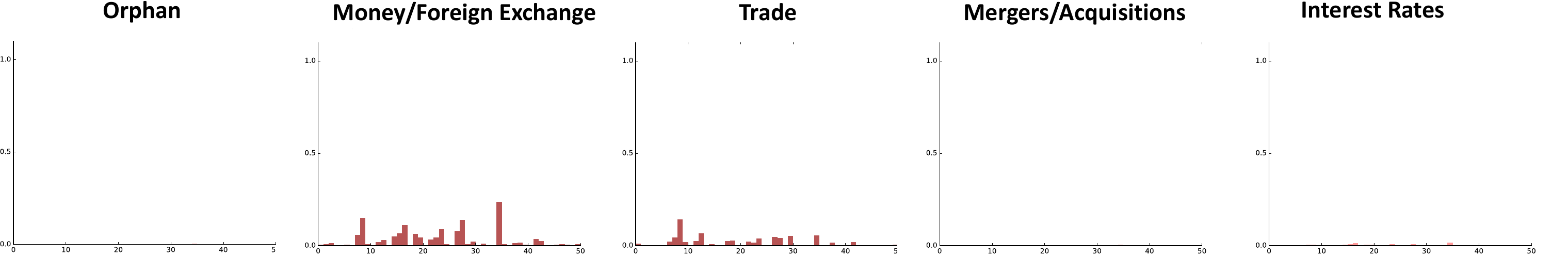} }\\
         \hline         
		\bottomrule
	\end{tabular}
	}
 \caption{Fully Corrected Case with weakly confidence ($ 0.4 <p< 0.6$). Although category-specific phrases are mentioned only once, category labels are correctly predicted.}   \label{tab:a-1}
\end{table}
\begin{table}[h]
	\centering
	\resizebox{1\columnwidth}{!}{
	\begin{tabular}{m{15cm}|c|c}
		\toprule
        \hline
		 \textbf{PAPERS SAY VENEZUELAN CENTRAL BANK CHIEF TO RESIGN} & \textbf{Interest Rates}&\textbf{Money/Foreign Exchange}\\
         \hline 
  Venezuelan Central Bank President Hernan
  Anzola has submitted his resignation and asked President Jaime
  Lusinchi to transfer him to a post in the oil industry, two
  leading Venezuelan newspapers reported.
      The El Universal and El Nacional papers said Anzola would
  leave his position soon. Lusinchi already has decided on his
  successor, the El Nacional reported.
      Central Bank officials were not available for comment.
      Banking sources said Anzola differed with the Finance
  Ministry over economic policy, particularly over the direction
  of \textcolor{red}{interest rates}. He favoured raising the rates, which are
  currently well below the annual inflation rate of 33.2 pct.
      But the sources said he ran into opposition from Finance
  Ministry and government officials who thought an interest
  increase would fuel inflation.
	&\makecell{\\ \includegraphics[width=3.0in]{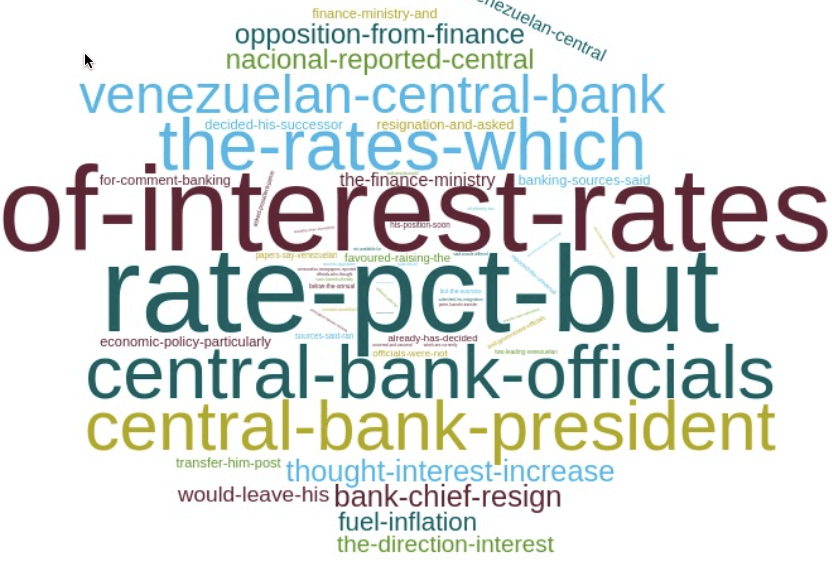} }
	&\makecell{\\ \includegraphics[width=3.5in]{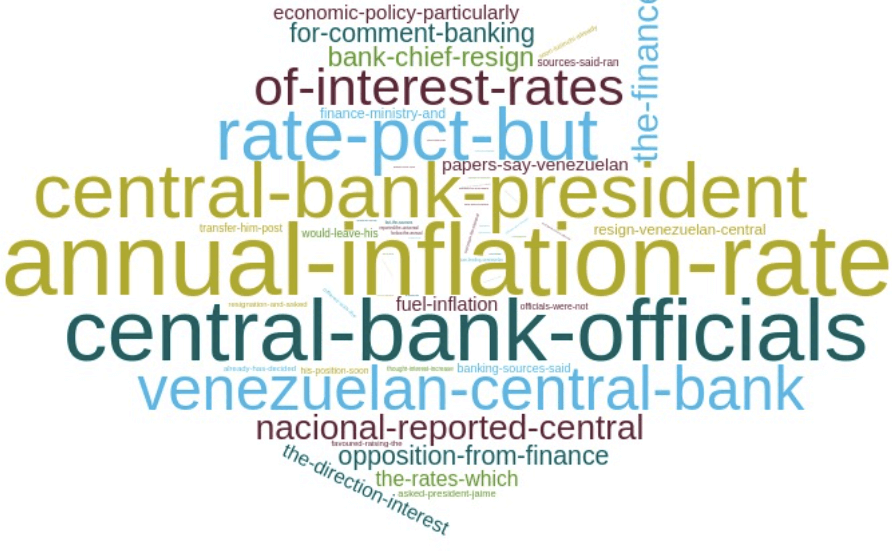} }
    \\
         \bottomrule        
         \multicolumn{3}{l}\makecell{\includegraphics[width=13in]{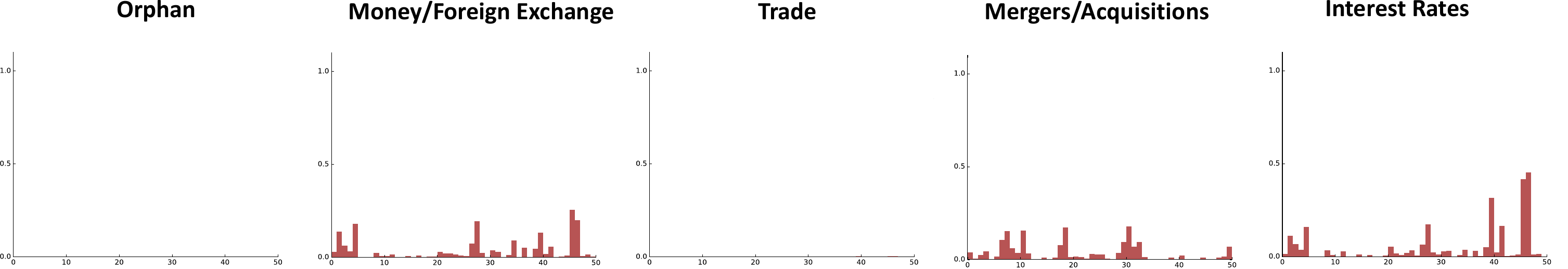} }\\
         \hline           
		\bottomrule
	\end{tabular}
	}
    \caption{Partially Corrected Case. The label of "Interest Rates" is correctly predicted but "Money/Foreign Exchange" is confused with "Mergers/Acquisitions" since the category-specific phrases are subtle and not be mentioned directly.} \label{tab:a-2}
\end{table}


\begin{table}[h]
	\centering
	\resizebox{0.9\columnwidth}{!}{
	\begin{tabular}{|c|c|c|c|c|c|c|c|c}
		\toprule
		 \textbf{Index}& \textbf{Routing} &\textbf{Leaky} &\textbf{Shared}& \textbf{OrphanCap} & \textbf{Loss}  & \textbf{Squash Coefficient} & \textbf{Accuracy}\\
		\midrule
	1	& 1 & Yes &Yes &  No &  Margin  & $|x|/(1+|x|)$ &  80.4  \\
	2	& 5 & Yes &Yes &  No &  Margin  & $|x|/(1+|x|)$ &  81.1 \\
	3	& 2 & Yes & Yes &  No &  Margin & $|x|/(1+|x|)$ &  80.5 \\
	4	& 3 & Yes & No &  Yes &  Margin  & $|x|/(1+|x|)$ & \textbf{82.3} \\
	5	& 3 & Yes & Yes &  Yes &  Margin  & $|x|/(1+|x|)$ &  81.8 \\
    6	& 3 & Yes & No &  No &  Margin  & $|x|/(1+|x|)$ &  81.9 \\
    7	& 3 & Yes & Yes &  No &  Margin  & $|x|/(1+|x|)$ &  81.2 \\
    8	& 3 & No & Yes &  No &  Margin  & $|x|/(1+|x|)$ &  80.9 \\
    9	& 3 & Yes &Yes &  No &  Margin  & $|x|/(1+|x|)$&  81.6 \\
	10	& 3 & Yes &Yes &  No &  Spread  & $|x|/(1+|x|)$&  81.1 \\
	11	& 3 & Yes  & Yes &  No &  CrossEnt  &$|x|/(1+|x|)$&  80.3 \\
	12	& 3 & Yes & Yes &  No &  Margin  & $1-\exp(-|x|)$&  80.5 \\
	13	& 3 & Yes & Yes &  No &  Margin  & $\text{tanh}(|x|)$&  80.8  \\
	14	& 3 & Yes & Yes &  No &  Margin  & $\text{None}$ &  80.6 \\
                \midrule		
		\bottomrule
	\end{tabular}
	}
    \caption{The effect of varying different components of Capsule-B on MR dataset. ``Routing'': represent the number of the routing iteration. ``Leaky'': use leaky softmax or not. ``Shared'': use shared weights between child-parent relationships or not. ``OrphanCap'': use orphan category or not.
    }
    \label{tab:a-4}
    \vspace{-0.5cm}
\end{table}

\end{document}